\documentclass[runningheads]{llncs}
\makeatletter
\providecommand{\@LN@col}[1]{}
\providecommand{\@LN}[2]{}
\makeatother
\usepackage[T1]{fontenc}
\usepackage{newtxtext}
\usepackage[varvw]{newtxmath}
\usepackage{graphicx}

\usepackage{array}
\usepackage{algorithm}
\usepackage{algpseudocode}
\usepackage{placeins}
\usepackage{cite}
\usepackage[hidelinks]{hyperref}
\usepackage{todonotes}
\newcommand{\datasetURL}{\url{https://huggingface.co/datasets/theshivam7/bharatsynth-qa}}

\title{BharatSynthQA: A Context-Grounded Question Answering Dataset on Indian Temples and Religious Heritage}
\titlerunning{BharatSynthQA}
\author{Riya Naik\inst{1}\orcidID{0009-0009-3826-4443}\and
Shivam Sharma \inst{2} \and
Swati Agarwal\inst{3}}
\authorrunning{R Naik et al.}
\institute{BITS Pilani, K K Birla Goa Campus, India \\
\email{p20210056@goa.bits-pilani.ac.in} \\ 
\and
Indian Institute of Technology Madras, India\\
\email{22f3002256@ds.study.iitm.ac.in} \\
\and
PandaByte Innovations Pvt Ltd, India\\
\email{agr.swati90@gmail.com}}

\begin{document}
\maketitle
\begin{abstract}
India's temples and other places of worship are among its richest repositories of cultural heritage, bringing together religious tradition, history, architecture, and art. Due to the widespread curiosity, people increasingly turn to Artificial Intelligence (AI) assistants for information about them. Current models are rarely trained or evaluated on this domain. We present \textit{BharatSynthQA}, a passage grounded question answering (QA) dataset focused on Indian temples, which also covers other religious sites connected to them. Our pipeline crawls Wikipedia from topics seeded by authentic heritage sources, filters articles for topical relevance, and locations in India. Further it generates Boolean, factoid, and explanatory QA pairs with a fewshot prompted LLM. An LLM judge filters the pairs on answer correctness, retaining 98,677 of 131,494 generated pairs. Additionally, 300 human-annotated pairs are released separately as a reference set. Human evaluation of the subsample of this dataset showed 89.83\% of the questions were formed correctly. Benchmarking five current LLMs shows that factoid token F1 is only 0.18--0.30 without the source passage but 0.73--0.77 with it, indicating that this domain is poorly captured in model knowledge. The dataset is available at \datasetURL{} to support research on culturally grounded QA and retrieval-augmented generation for Indian heritage.
\end{abstract}

\section{Introduction}
India is home to one of the oldest civilizations, and its places of worship are central to that heritage. Temples, along with other religious sites, are also records of history, architecture, art, and regional tradition. Interest in this heritage has grown both domestically and globally, and people increasingly turn to digital assistants for information about it. However, the data used to train and evaluate these systems contains little about Indian temples and other religious sites. Answering such queries reliably also requires more than factual lookup: it often calls for explanations that draw on interconnected historical and regional context. Yet no large scale QA dataset exists to train or evaluate models on this kind of culturally grounded question answering. Therefore, a dedicated data set is needed both to measure how well current LLMs understand this domain and to support culturally aware applications in education, tourism, and digital heritage preservation.

\noindent
The distinction between a traditional QA system and a Machine Reading Comprehension QA (MRCQA) system is that the latter is designed to comprehend the context and answer questions based directly on the information within the context \cite{varghese2023question}. The backbone of MRCQA is the robust datasets, such as MRQA \cite{fisch2019mrqa} and SQuAD \cite{rajpurkar2018know}, that contain questions paired with passages. But they come with trade-offs, as their creation requires domain expertise, making annotation time-consuming and costly. And thus many domains suffer from a lack of sufficient quality data to develop robust QA datasets. Due to their manual nature, they often fail to represent complex reasoning scenarios reliably \cite{varghese2022comprehensive}.

\noindent
MRCQA datasets often focus on open or popular domains such as medicine or law, and few widely recognized datasets spans historical texts, sites, traditions, and languages. Which touch these topics only indirectly, such as, WikiQA \cite{yang2015wikiqa} or TyDi QA \cite{clark2020tydi}. 

\noindent
Current LLMs can now synthesize training and evaluation data at far lower effort than human annotation, which makes them a practical way to build datasets for niche or under-represented domains (Section~\ref{sec:related}). In this paper, we propose a pipeline that uses LLMs to generate BharatSynthQA, a context-grounded QA dataset focused on Indian temples, which also spans other religious sites connected to them.

\noindent
Specifically, we aim to answer the following research questions:
\begin{itemize}
    \item RQ1: Can topic-based crawling, combined with filtering, collect a corpus of domain-relevant documents on Indian temples and religious heritage?
    \item RQ2: How reliably can a prompted LLM, combined with LLM-judge, generate passage-grounded QA pairs on Indian temples and religious heritage across diverse question types?
    \item RQ3: How well do current LLMs answer questions on Indian temples and religious heritage from their own knowledge, and how much does access to the source passage help?
\end{itemize}

\section{Methodology}
Our pipeline first collects Wikipedia documents on Indian temples and other religious sites based on authentic seed documents, and then generates accurate QA pairs from them, ranging from simple factoid to explanatory questions.

\subsection{Materials}

\textbf{Data:} To build the corpus, we first selected a set of seed documents. Each seed had to fall within the scope of the dataset and come from an authentic source, such as an official temple trust website or a government tourism portal. Examples include the Somnath Temple Trust\footnote{\url{https://somnath.org/somnath-darshan/}} and Uttarakhand Tourism's Char Dham Data \footnote{\url{https://www.uttarakhandtourism.gov.in/experiences/char-dham}}. We extracted topics from these seed documents. These topics then guided the crawler towards relevant documents (Section~\ref{sec:doc_collection}).

\noindent
\textbf{Models and Frameworks:} We use several LLMs, through the Together AI and OpenAI APIs, in three roles. Llama 4 Scout\footnote{\url{https://www.together.ai/models/llama-4-scout}} generates the QA pairs, and DeepSeek V4 Flash 0731\footnote{\url{https://www.together.ai/models/deepseek-v4-flash-0731}} judges the quality of the generated pairs (LLM-as-a-judge). The models used to evaluate performance on BharatSynthQA are described in Section~\ref{sec:model_evals}. All implementations are in Python 3.10.

\subsection{Document Collection and Analysis}
\label{sec:doc_collection}
Using the topics extracted from the seed documents, we implemented a topic-based crawler over Wikipedia. We chose Wikipedia as the starting point for collection because its articles are citation-backed and verifiable, which makes them more reliable web sources. Algorithm~\ref{alg:cap} describes the crawler. Seeded from temple sources, the crawler also reached other religious sites through linked topics such as history and pilgrimage. We preserved them as part of India's shared religious heritage, so the corpus is temple-centered but not temple-only. Alongside the crawled articles, we extracted location data from OpenStreetMap (OSM)\footnote{OpenStreetMap = \url{https://www.openstreetmap.org/}}.

\begin{algorithm}[tb]
\caption{Topic-Based Crawler}\label{alg:cap}
\textbf{Input}: $S$: Seed Articles, $W$: WikiArticles, $\tau$: relevance threshold\\
\textbf{Output}: a set $R$ = Relevant Documents; $L$: Less Relevant Documents
\begin{algorithmic}[1]
\Procedure{TopicBasedCrawler}{$S, W, \tau$}
    \State $ST \gets$ getSeedTopics($S$)
    \State $R \gets [], L \gets []$
    \For{$w$ in $W$}
        \State $WT \gets$ getWikiTopics($w$)
        \If{similarityScore($WT, ST$) $\geq \tau$}
            \State $R$.append($w$)
            \State saveContent($w$)
        \Else
            \State $L$.append($w$)
            \State saveContent($w$)
        \EndIf
    \EndFor
    \For{$link$ in $R \cup L$}
        \State $W$.push($link, similarityScore$)
    \EndFor
    \State \textbf{return} $R, L$
\EndProcedure
\end{algorithmic}
\end{algorithm}

\begin{figure}[tb]
    \centering
    \includegraphics[draft=false]{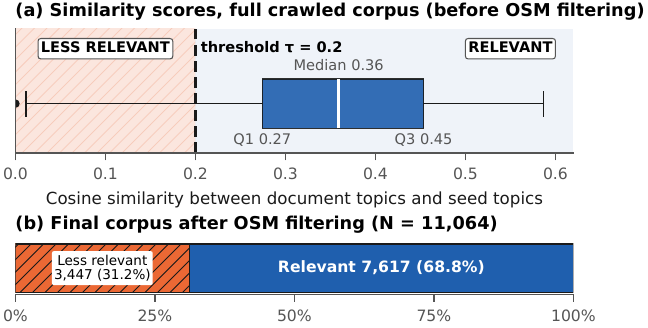}
    \Description{Two panels. Panel (a) is a horizontal box plot of topic cosine similarity scores for the full crawled corpus before OSM filtering, ranging from 0 to 0.60 with a median of 0.36. A dashed line at 0.2 marks the relevance threshold. Panel (b) shows the final corpus of 11,064 documents after OSM filtering: 7,617 (68.8\%) relevant and 3,447 (31.2\%) less relevant.}
    \caption{(a) Distribution of topic cosine similarity scores across the full crawled corpus, before OSM filtering. The dashed line marks the relevance threshold $\tau = 0.2$. (b) Relevance split of the final corpus after OSM filtering.}
    \label{fig:threshold}
\end{figure}

\noindent
We extracted the topics of each collected document using the TextRazor API\footnote{TextRazor = \url{https://www.textrazor.com/}} and compared them with the seed topics using cosine similarity. Before OSM filtering, the similarity scores ranged to 0.60 (Figure~\ref{fig:threshold}a). Based on this distribution, we set a relevance threshold of $\tau = 0.2$. Documents scoring at or above $\tau$ were labeled \textit{relevant}, and the rest \textit{less relevant}. Although not used in QA generation, we did not discard the less relevant documents because they were not outright false positives. They carried information adjacent to the relevant documents and can serve as supporting context for future exploration.

\noindent
Further, we apply OSM-based geographic filtering to exclude heritage sites outside India. Using OSM data and the Google Geocoding API\footnote{Google Geocoding = \url{https://developers.google.com/maps/documentation/geocoding}}, we geocode articles that describe a physical site and remove those located outside India. Articles not tied to a single location, such as those on traditions, philosophies, or languages, are retained. We post-process the documents to extract images and discard unreferenced text. After filtering, we segment 11,064 documents: 7,617 relevant and 3,447 less relevant (Figure~\ref{fig:threshold}b).
The share of less relevant documents rises in the final corpus. This is because the OSM filter removed some documents with heritage sites outside India such as monuments in neighboring countries that share traditions with India. They matched the domain topics closely but fell outside our geographic scope.

\subsubsection{Document Analysis}
\begin{figure}[tb]
    \centering
    \includegraphics{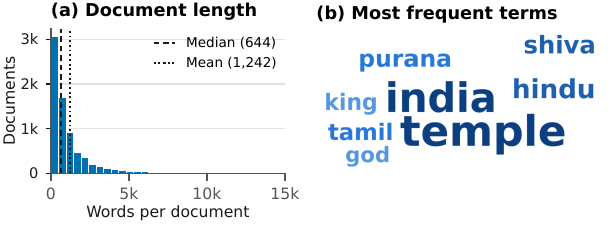}
    \Description{Two panels. (a) A right-skewed histogram of document lengths in words, with most documents under 2,000 words; dashed and dotted lines mark the median of 644 and the mean of 1,242 words. (b) A word cloud of the most frequent terms, led by temple and india, followed by hindu, shiva, purana, tamil, king, and god.}
    \caption{Source documents: (a) document length  with the median and mean marked; (b) most frequent terms after lemmatization and sized by token count.}
    \label{fig:corpus_eda}
\end{figure}

We characterize the source documents using the 7,617 relevant Wikipedia documents (Figure~\ref{fig:corpus_eda}). Document lengths are right-skewed, with a median of 644 words and a mean of 1,242 words. As the scraping intended the most frequent processed terms are temple, india, hindu, shiva, purana, and tamil. These highlights that the documents mainly represents temples, deities, and scriptures in India. 

\subsection{Question Answer pairs Generation}
Our goal here is to build a dataset with QA pairs about Indian temples and religious heritage, ranging from basic fact-based to explanatory questions. 
We first gather QA pairs through human annotation and then use them to generate further QA pairs with an LLM, with all answers grounded in Wikipedia passages.

\subsubsection{Human Annotation}
\label{sec:humman_annotation}
We approached ten doctoral and postgraduate students with experience in machine learning, text analytics, and mining. Each annotator was provided with around ten document contexts and a detailed explanation of the questions to be asked. Annotators were guided to create various types of QA pairs categorized into the following categories:

\begin{itemize}
    \item Boolean: Questions that can be fully and correctly answered with ``Yes'' or ``No.''
    \item Factoid: Questions seeking one specific fact, such as a name, date, number, place, or short phrase.
    \item Other: Questions seeking an explanation, process, reason, method, comparison, or description that requires several connected statements.
\end{itemize}

\begin{figure}
    \centering
    \includegraphics[draft=false]{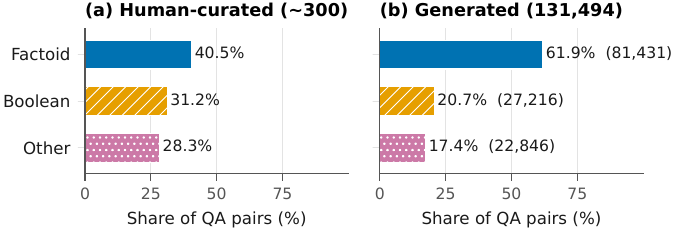}
    \Description{Two bar charts of question-type shares. Human-curated set: Factoid 40.5, Boolean 31.2, Other 28.3 percent. Generated set: Factoid 61.9 percent (81,431 pairs), Boolean 20.7 percent (27,216), Other 17.4 percent (22,846).}
    \caption{Question-type distribution in (a) the human-curated annotation set and (b) the full generated BharatSynthQA dataset (131,494 QA pairs)}
    \label{fig:qa_types}
\end{figure}

\noindent
Question types are determined by the information requested in the question, independently of the provided answer's length, style, or quality. Annotation used LabelStudio\footnote{LabelStudio = \url{https://labelstud.io/}}. With this approach, we collected approximately 300 QA pairs across the categories. These pairs supply the few-shot examples for LLM QA generation. The distribution of collected QA pairs is depicted in Figure~\ref{fig:qa_types}a.

\subsubsection{Prompt-based Question Generation}
Annotating the full corpus manually was impractical, so we use few-shot prompting to generate QA pairs from the 7,617 relevant documents. The prompt asks the LLM for QA pairs that resemble the human-annotated examples and follow the three question types of Section~\ref{sec:humman_annotation} human annotated pairs supply the few-shot examples.

\noindent
We compared three candidate generators Llama 3.2 90B, Llama 4 Scout and DeepSeek on sub sample of ten documents. We selected Llama 4 Scout at temperature 0.7 because it provided (145) unique questions and balanced distribution across question types.

\noindent
Each document was segmented to fit Llama 4 Scout's input budget together with the few shot prompt, and sent to the model through LangChain.. 
As a result, our method produced 131,494 QA pairs (Figure~\ref{fig:qa_types}b), from simple fact-based questions to explanatory questions that draw on several connected statements in the passage. Table~\ref{tab:examples} shows an example passage with one generated QA pair of each type.

\begin{table}
\caption{Example from BharatSynthQA: a passage excerpt with one synthetic QA pair of each question type from the dataset.}\label{tab:examples}
\centering
\scriptsize
\setlength{\tabcolsep}{4pt}
\begin{tabular}{@{}l >{\raggedright\arraybackslash}p{5.2cm} >{\raggedright\arraybackslash}p{4.6cm}@{}}
\hline
\multicolumn{3}{@{}p{11.0cm}@{}}{\textbf{Passage (excerpt):} Kottekkad Temple (Wikipedia). Kottekkad Temple is a Hindu temple situated about 8 km from Thrissur, Kerala, India. It has pratishtas of Bhagavathi, Shiva, Aanganear (Hanuman), Ayyappan, Nagathanmar, Subhramaniyan (Murugan), and Bhrahmarakshassu.} \\
\hline
Type & Question & Answer \\
\hline
Boolean & Is Kottekkad Temple a Buddhist temple? & No \\
Factoid & In which state is Kottekkad Temple located? & Kerala \\
Other & Can you describe Kottekkad Temple? & Kottekkad Temple is a Hindu temple situated about 8 km from Thrissur, Kerala, India. It has pratishtas of Bhagavathi, Shiva, Aanganear (Hanuman), Ayyappan, Nagathanmar, Subhramaniyan (Murugan), and Bhrahmarakshassu. \\
\hline
\end{tabular}
\end{table}

\begin{figure}
    \centering
    \includegraphics[width=\linewidth]{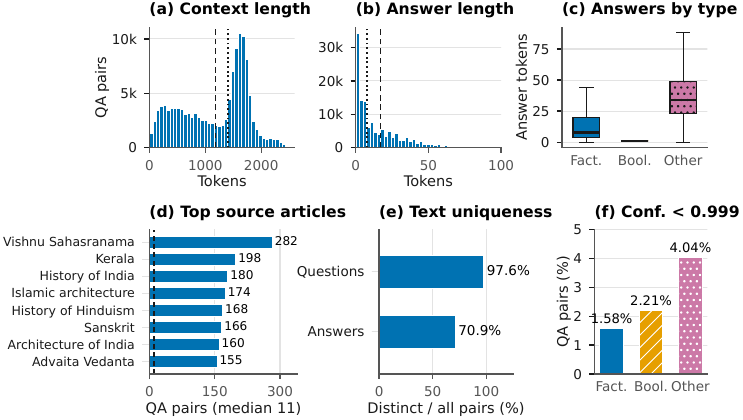}
    \caption{Profile of the 131,494 generated QA pairs. Top: token lengths of (a)~contexts and (b)~answers, with dashed (mean) and dotted (median) lines, and (c)~answer length by question type. Bottom: (d)~the eight most-represented source articles; (e)~distinct normalized strings as a share of all pairs; (f)~share of pairs with generation confidence below 0.99.}
    \label{fig:gen_profile}
\end{figure}

\noindent
As can be seen from Figure~\ref{fig:gen_profile} Boolean answers are a single token, Factoid answers have a median of 8 tokens, and explanatory \textit{Other} answers a median of 34 tokens.
Pairs are spread across many source articles, with a median of 11 per article (Figure~\ref{fig:gen_profile}d). Questions are highly diverse around 97.6\% of question strings are distinct as shown in Figure~\ref{fig:gen_profile}e). While generating QA pairs the model was also instructed to set a confidence score for how sure the model is while generating that respective QA pair. Few pairs fall below a generation confidence of 0.99 (Figure~\ref{fig:gen_profile}f). The generator is therefore least certain on explanatory questions, consistent with their lower LLM-judge retention (Section~\ref{sec:judge}).

\section{Dataset Evaluation}
We evaluated our dataset in two fold: human evaluated sub sample of the dataset and then judge model uses these as a reference to evaluate the dataset overall.

\subsection{Human Evaluation}
\label{sec:human_eval}
\begin{figure}
    \centering
    \includegraphics{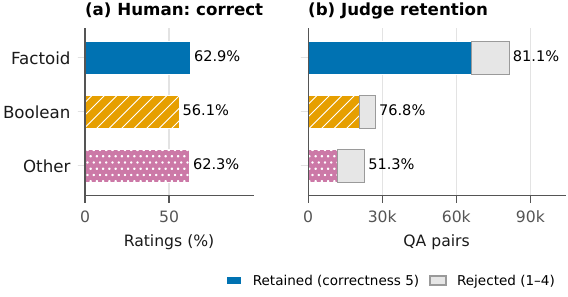}
    \caption{Quality assessment by question type. (a)~Human evaluation: share of ratings giving the answer the highest correctness rating (Correct), over 600 ratings on 200 QA pairs, each reviewed by three evaluators. (b)~LLM-judge filtering of all 131,494 pairs: pairs are retained only with overall correctness 5 and rejected otherwise (scores 1--4).}
    \label{fig:quality}
\end{figure}

Three human evaluators each reviewed the same 200 QA pairs, sampled from the generated pairs before judge filtering: 133 Factoid, 44 Boolean, and 23 Other pairs, giving 600 ratings in total. Evaluators assessed four criteria: question clarity, answer correctness, whether the supplied passage was needed to answer the question, and question-type correctness. Table~\ref{tab:human_evaluation} summarizes the outcomes. It showed that ~90\% of the formed questions were clear and if we take full and partial correctness then 84.16\% answers were correct. This implied that pipeline generated valid context grounded QA pairs.
Figure~\ref{fig:quality}a reports, per question type, the share of ratings that gave the highest rating: 62.91\% for Factoid, 56.06\% for Boolean, and 62.32\% for Other pairs.

\begin{table}
\caption{Human-evaluation outcomes on 200 distinct LLM generated QA pairs.}\label{tab:human_evaluation}
\centering
\begin{tabular}{lr}
\hline
Assessment outcome & Percentage \\
\hline
Question marked clear & 89.83\% \\
Answer marked fully correct & 61.33\% \\
Answer marked partially correct & 22.83\% \\
Answer marked incorrect & 15.83\% \\
Passage marked necessary & 68.83\% \\
\hline
\end{tabular}
\end{table}

\subsection{LLM as a Judge}
\label{sec:judge}

We evaluated all 131,494 QA pairs using DeepSeek V4 Flash 0731 as our LLM judge, with temperature 0 and high reasoning effort. We selected the judge for its agreement with human judgment and its cost efficiency. To compare candidates, we ran four representative LLM judges on pairs drawn from the 200 human-evaluated pairs (Section~\ref{sec:human_eval}), using a common rubric and temperature 0 (Table~\ref{tab:judge_selection_pilot}). All four agreed closely with the human ratings (89--92\%). GPT-5.4 Nano agreed most closely (92.0\%) but cost almost three times as much per 1,000 pairs as DeepSeek V4 Flash 0731, which reached 91.0\% at the lowest cost (USD 0.325). DeepSeek V4 Flash 0731 therefore offered the best balance of agreement and cost for judging the full dataset.

\noindent
Each pair was assessed against its source context on seven dimensions, scored from 1 to 5: question clarity, answer correctness, context relevance, type correctness, faithfulness, reasoning validity, and answer completeness. Table~\ref{tab:judge_quality} reports the scores before filtering. We retained pairs with an answer-correctness score of 5. This yields 98,677 pairs (75.0\%): 66,055 Factoid, 20,906 Boolean, and 11,716 Other pairs (Figure~\ref{fig:quality}b). A stratified split of the retained pairs by question type, with random seed 42, gives 69,074 training, 9,868 validation, and 19,735 test pairs.

\begin{table}
\caption{Comparison of representative LLM judges on a 20-pair pilot drawn from the human-evaluated pairs. Bold identifies the selected model.}
\label{tab:judge_selection_pilot}
\centering
\begingroup
\setlength{\tabcolsep}{3pt}
\renewcommand{\arraystretch}{1.15}
\begin{tabular}{@{}lcrrr@{}}
\hline
\textbf{Judge model} & \textbf{Reasoning} & \textbf{Human} & \textbf{Latency} & \textbf{Cost/1,000} \\
 & \textbf{effort} & \textbf{agreement (\%)} & \textbf{(mean s)} & \textbf{(USD)} \\
\hline
\textbf{DeepSeek V4 Flash 0731} & High & 91.0 & 5.56 & \textbf{0.325} \\
GLM-5.3 Flash & High & 90.0 & 4.47 & 0.440 \\
GPT-5.4 Nano & Medium & 92.0 & 4.37 & 0.915 \\
Gemini 3.6 Flash & Medium$^{a}$ & 89.0 & 4.75 & 8.908 \\
\hline
\end{tabular}
\endgroup
\par\smallskip
\end{table}

\begin{table}
\caption{Mean LLM-judge scores before filtering (1--5), grouped by assigned question type.}\label{tab:judge_quality}
\centering
\begin{tabular}{lrrr}
\hline
Dimension & Factoid & Boolean & Other \\
\hline
Question clarity & 4.32 & 4.20 & 4.01 \\
Answer correctness & 4.30 & 4.08 & 3.62 \\
Context relevance & 4.41 & 4.18 & 4.13 \\
Type correctness & 4.44 & 4.25 & 4.13 \\
Faithfulness & 4.32 & 4.08 & 3.89 \\
Reasoning validity & 2.16 & 2.20 & 2.78 \\
Answer completeness & 4.28 & 4.08 & 3.31 \\
\hline
\end{tabular}
\end{table}

\subsection{Model Performance Evaluation}
\label{sec:model_evals}

We benchmarked five QA-answering models: DeepSeek V4.1 Flash\footnote{\url{https://www.together.ai/models/deepseek-v4-1-flash}}, GLM-5.3\footnote{\url{https://www.together.ai/models/glm-5-3}}, GPT-5.6 Luna\footnote{\url{https://developers.openai.com/api/docs/models/gpt-5.6-luna}}, GPT-6 Luna\footnote{\url{https://developers.openai.com/api/docs/models/gpt-6-luna}}, and Muse Glimmer 30B\footnote{\url{https://www.together.ai/models/muse-glimmer}} under two zero-shot conditions: answering without context and answering with the source passage. We sampled 616 pairs of each type, giving 1,848 evaluation questions. All ten model condition runs used the same question identifiers and completed successfully.
We report Boolean balanced accuracy, Factoid token F1, and Other ROUGE-L separately (Table~\ref{tab:benchmark}). We observed that providing the passage improves all three reported metrics for every model (Table~\ref{tab:benchmark}; see Section~\ref{sec:discussion}).
Without the passage, current LLMs struggle with this content: Factoid token F1 is only 0.177--0.297 and Other ROUGE-L 0.166--0.216, and Boolean balanced accuracy ranges from 0.62 to 0.81. With the passage, scores rise to 0.725--0.766, 0.459--0.596, and 0.93--0.98 respectively. Knowledge of Indian temples and religious heritage is thus poorly captured in model parameters, and the grounded passages supply what models lack. We believe that training on BharatSynthQA will provide future possibility of closing this gap. 

\begin{table}
\caption{Zero-shot QA answering without (No) and with (Psg.) the supplied source passage on the same stratified subset: Boolean balanced accuracy, Factoid token F1, and Other ROUGE-L. Bold marks the best score in each column.}\label{tab:benchmark}
\centering
\setlength{\tabcolsep}{5pt}
\begin{tabular}{lcccccc}
\hline
 & \multicolumn{2}{c}{Boolean bal.\ acc.} & \multicolumn{2}{c}{Factoid token F1} & \multicolumn{2}{c}{Other ROUGE-L} \\
\cline{2-3}\cline{4-5}\cline{6-7}
Model & No & Psg. & No & Psg. & No & Psg. \\
\hline
DeepSeek V4.1 Flash & \textbf{0.81} & 0.96 & \textbf{0.30} & 0.75 & 0.21 & 0.59 \\
GLM-5.3 & 0.74 & 0.97 & 0.26 & \textbf{0.77} & 0.19 & 0.58 \\
GPT-5.6 Luna & 0.78 & 0.97 & 0.27 & 0.73 & \textbf{0.22} & 0.47 \\
GPT-6 Luna & 0.76 & \textbf{0.98} & 0.27 & 0.72 & 0.21 & 0.46 \\
Muse Glimmer 30B & 0.62 & 0.93 & 0.18 & 0.75 & 0.17 & \textbf{0.60} \\
\hline
\end{tabular}
\end{table}

\section{Discussion}
\label{sec:discussion}

\textbf{Collecting a domain-relevant corpus.} Topic-based crawling seeded from authentic temple sources, followed by topic-similarity and geographic filtering, produced a largely on-domain corpus. Of the 11,064 documents retained after OSM filtering, 7,617 (68.8\%) cleared the relevance threshold of $\tau = 0.2$ and form the generation corpus (Figure~\ref{fig:threshold}). The most frequent terms (temple, hindu, shiva, purana) confirm that the corpus centres on temples, deities, and scriptures, while linked topics such as history and pilgrimage brought in other religious sites that share these traditions. Geographic filtering proved essential, because many documents with high topic similarity described sites in neighbouring countries.

\noindent
\textbf{Reliability of generated QA pairs.} Few-shot prompting with Llama 4 Scout produced 131,494 QA pairs spanning all three question types, with 97.6\% distinct questions and answer lengths that separate the types as intended (Figure~\ref{fig:gen_profile}c). Human evaluation of a 200-pair sample showed that the raw generations are largely usable but not error-free: 61.3\% of ratings marked the answer fully correct, 22.8\% partially correct, and 15.8\% incorrect. Using these human judgments as a reference, the LLM judge filtered the full set and retained 98,677 pairs which scored fully correct. Explanatory questions are the hardest to generate: they have the lowest judge retention (51.3\%) and the highest share of low-confidence generations (4.04\%).

\noindent
\textbf{Model knowledge vs.\ passage access.} All five models answer poorly from their own knowledge but well once given the passage (Table~\ref{tab:benchmark}): Factoid token F1 rises from at most 0.30 to at least 0.72, and Boolean balanced accuracy from 0.62--0.81 to 0.93--0.98. Knowledge of Indian temples and religious heritage is therefore poorly captured in model parameters, while the models read the supplied passages effectively. Passage access also matters more than the choice of model: Muse Glimmer 30B, the weakest model without the passage, achieves the best \textit{Other} ROUGE-L with it. Explanatory questions remain the most challenging even with context (ROUGE-L at most 0.60), although ROUGE-L also under scores correct answers that paraphrase the reference. BharatSynthQA thus serves both as a closed-book probe of cultural knowledge and as a benchmark for reading comprehension and retrieval-augmented QA.

\section{Related work}
\label{sec:related}

\textbf{Existing Datasets.}
Current MRCQA datasets range from multi-hop reasoning, such as HotpotQA \cite{yang2018hotpotqa}, where answers are synthesized from multiple documents, to multi-turn interactions \cite{yatskar2019qualitative} and domain-specific settings \cite{lamichhane2023enhancing}. Cross-lingual resources such as XQuAD \cite{artetxecross} extend SQuAD to other languages, and NativQA \cite{hasan2025nativqa} collects 63,757 manually annotated QA pairs in seven languages from nine regions, built from natural search queries on 18 everyday topics. Closer to our domain, DOSA \cite{seth2024dosa} gathers 615 social artifacts from 19 Indian geographic subcultures through participatory research, but does not pose QA pairs. SANSKRITI \cite{maji2025sanskriti} benchmarks LLM knowledge of Indian culture with 21,853 question--answer pairs spanning 28 states and 8 union territories, and MILU \cite{verma2024milu} draws 79,617 multiple-choice questions from Indian regional and state-level examinations in 11 languages, covering topics such as local history, arts, and festivals. Both probe what a model already knows: their questions come without a supporting passage. BharatSynthQA differs from these resources in three ways. First, every answer is grounded in a source passage, so the dataset supports reading-comprehension and retrieval-augmented evaluation and lets models be compared with and without context (Section~\ref{sec:model_evals}). Second, it is larger than existing India-focused resources: 131,494 generated QA pairs, of which 98,677 pass judge filtering, with approximately 300 human-annotated pairs released separately. Third, its questions are graded from Boolean and Factoid to explanatory \textit{Other} questions rather than following a single format. Its scope is also deliberately narrower and deeper: it centres on Indian temples, together with other religious sites connected to them, and the history, architecture, and traditions around them. Table~\ref{tab:dataset_comparison} summarizes these differences.

\begin{table}[tb]
\caption{Comparison of BharatSynthQA with related QA and cultural-knowledge datasets. Size counts QA pairs or questions (DOSA: artifacts).}\label{tab:dataset_comparison}
\centering
\scriptsize
\setlength{\tabcolsep}{4pt}
\begin{tabular}{@{}l r >{\raggedright\arraybackslash}p{3.4cm} l >{\raggedright\arraybackslash}p{2.9cm}@{}}
\hline
Dataset & Size & Domain & Lang. & Question types \\
\hline
SQuAD 2.0 \cite{rajpurkar2018know} & 150K & Open (Wikipedia) & En & Extractive; unanswerable \\
TyDi QA \cite{clark2020tydi} & 204K & Open (Wikipedia) & 11 & Extractive; Yes/No \\
WikiQA \cite{yang2015wikiqa} & 3K & Open (search queries) & En & Answer-sentence selection \\
NativQA \cite{hasan2025nativqa} & 64K & Everyday regional topics & 7 & Free-form \\
DOSA \cite{seth2024dosa} & 615 & Indian social artifacts & En & Not QA \\
SANSKRITI \cite{maji2025sanskriti} & 21.9K & Indian culture (36 states/UTs) & En & Multiple-choice \\
MILU \cite{verma2024milu} & 79.6K & India-centric exam subjects & 11 & Multiple-choice \\
\hline
\textbf{BharatSynthQA} & \textbf{98.7K} & Indian temples and religious heritage & En & Boolean, Factoid, explanatory \\
\hline
\end{tabular}
\end{table}

\noindent
\textbf{LLMs for Data Generation.}
LLMs are increasingly used in place of human annotators to generate and augment datasets \cite{meng2022generating,wang2022pinto}. LLMs have contributed to data synthesis across various applications, such as sentiment classification \cite{choi-etal-2024-unigen}, named entity recognition \cite{naraki2024augmenting}, and generating QA data \cite{artetxecross,schmidt2024prompting}. \cite{wu2024unigen} developed a unified pipeline for data generation using seed data, description or data schema, and user constraints to curate controlled datasets over the original ones. 
Synthetic data, though mainly used in generic domains \cite{abdullin2023synthetic}, has found its way into curating data for specific domains such as medical, legal, and even in the socio-cultural domains \cite{kumichev2024medsyn,namboori2023gemquad,abdallah2023exploring}. Recently, \cite{urailertprasert2024sea} constructed SEA-VQA, prompting LLMs to generate multiple-choice questions on images of Southeast Asian cultural traditions. Following this line of work, BharatSynthQA applies LLM-based generation to Indian temples and religious heritage, covering Boolean, factoid, and explanatory questions grounded in source passages.

\noindent
\textbf{Challenges in Curating Reliable Data.}
Reliable synthetic QA data must contend with ambiguous or unauthentic sources, model bias, and hallucinated answers, which calls for validation through cross-referencing, human review, and feedback \cite{kim2023aligning,sun2024principle,lee2023making}, transparent and reasoning-based generation \cite{wang2023let,narangself}, and robust evaluation such as LLM judges \cite{zheng2023judging}; our pipeline combines human evaluation with LLM-judge filtering for this reason.

\section{Conclusion}
We presented BharatSynthQA, a context-grounded question answering dataset centred on Indian temples that also covers other religious sites connected to them. Starting from authentic heritage sources, our pipeline crawls Wikipedia by topic similarity, filters documents for relevance and location in India, generates Boolean, factoid, and explanatory QA pairs with a few-shot prompted LLM, and filters them with an LLM judge. The resulting resource contains 98,677 judge-retained QA pairs with training, validation, and test splits, together with approximately 300 human-annotated pairs. Human evaluation shows that most questions are clear and grounded in their passages, and benchmarking five current LLMs reveals a large gap between answering from model knowledge and answering with the passage, confirming that this domain is under-represented in current models. BharatSynthQA supports several uses. As a benchmark, it can be used in a closed-book setting to probe how much LLMs know about Indian religious heritage, and with passages to evaluate reading comprehension and retrieval-augmented QA. As training data, its splits can be used to fine-tune or adapt models for culturally aware assistants in education, tourism and pilgrimage, and digital heritage preservation. The human-annotated pairs provide a gold reference set for validating future generation and evaluation methods, and the pipeline itself can be reused to build grounded QA resources for other regions and heritage domains. The dataset is available at \datasetURL.

\section*{Future Directions}
BharatSynthQA is temple-centred: Hindu temples and religious content dominate, and other traditions and regions are covered less. Broadening the seed sources and capping the number of pairs per source article would yield a more balanced resource. All passages come from English Wikipedia, so the dataset is English and Wikipedia only extending it to Indian languages and to additional authentic sources would widen its reach. The QA pairs are LLM-generated and filtered by a single LLM judge. A human audit of retained pairs with inter-annotator agreement, together with multiple LLM judges, would give stronger quality guarantees. The generated set is also Factoid-heavy,targeted generation of Boolean and explanatory questions would address the balance. Finally, the benchmark supplies the gold passage rather than retrieving it: a retrieval setting over the full collection of 11,064 documents, with the less relevant documents as realistic distractors, would enable end-to-end evaluation of retrieval-augmented QA.

\bibliographystyle{splncs04}
\bibliography{custom}

\end{document}